\documentclass{article} 
\usepackage{nips14submit_e,times}
\usepackage{hyperref}
\usepackage{url}
\usepackage[numbers,sort]{natbib}
\usepackage{graphicx}
\usepackage[aboveskip=4pt]{caption}
\usepackage[aboveskip=1pt]{subcaption}
\usepackage{amsmath,amssymb}
\usepackage{array}
\usepackage{multirow}
\usepackage{xcolor}
\usepackage{xspace}
\usepackage{nicefrac}
\usepackage[compact]{titlesec}

\titlespacing{\section}{0pt}{0pt}{0pt}
\titlespacing{\subsection}{0pt}{0pt}{0pt}

\title{Two-Stream Convolutional Networks \\ for Action Recognition in Videos}

\author{
Karen Simonyan \\
\And
Andrew Zisserman
}

\nipsfinalcopy 

\newcommand{\figref}[1]{Fig.~\ref{#1}}
\newcommand{\tblref}[1]{Table~\ref{#1}}
\newcommand{\sref}[1]{Sect.~\ref{#1}}

\newcommand*{\eg}{e.g.\@\xspace}
\newcommand*{\ie}{i.e.\@\xspace}

\DeclareMathOperator{\Rbb}{\mathbb{R}}
\DeclareMathOperator{\dvec}{\mathrm{\textbf{d}}}
\DeclareMathOperator{\pvec}{\mathrm{\textbf{p}}}

\begin{document}

\maketitle

\vspace{-3.5em}
\begin{center}
Visual Geometry Group, University of Oxford\\
\texttt{\{karen,az\}@robots.ox.ac.uk}  
\end{center}
\vspace{1em}

\begin{abstract}
We investigate architectures of discriminatively trained deep
Convolutional Networks (ConvNets) for action recognition in video. The
challenge is to capture the complementary information on appearance
from still frames and motion between frames. 
We also aim to generalise the best performing hand-crafted features within a data-driven learning framework.

Our contribution is three-fold. First, we propose a two-stream
ConvNet architecture which incorporates spatial and temporal networks.
Second, we demonstrate that a ConvNet trained on multi-frame dense
optical flow is able to achieve very good performance in spite of
limited training data.  Finally, we show that multi-task learning,
applied to two different action classification datasets, can be used
to increase the amount of training data and improve the
performance on both.

Our architecture is trained and evaluated on the standard video actions
benchmarks of UCF-101 and HMDB-51, where it is competitive with the state of the art.
It also exceeds by a large margin previous attempts to use deep nets for video classification.
\end{abstract}

\section{Introduction}
\label{sec:intro}
Recognition of human actions in videos is a challenging task which has
received a significant amount of attention in the research
community~\cite{Laptev08,Jhuang07,Wang13b,Karpathy14}.  Compared to
still image classification, 
the temporal component of videos
provides an additional (and important) clue for recognition, as a
number of actions can be reliably recognised based on the motion
information. Additionally, video provides natural data augmentation
(jittering) for single image (video frame) classification.

In this work, we aim at extending deep Convolutional Networks
(ConvNets)~\cite{Lecun89}, a state-of-the-art still image
representation~\cite{Krizhevsky12}, to action recognition in video data. This task
has
recently been  addressed in~\cite{Karpathy14} by using stacked video frames
as input to the network, but the results were significantly worse than
those of the best hand-crafted shallow
representations~\cite{Wang13b,Peng14}.  We investigate a different architecture
based on two separate recognition streams (spatial and
temporal), which are then combined by late fusion.  The spatial stream
performs action recognition from still video frames, whilst the
temporal stream is trained to recognise action from motion in the form
of dense optical flow.  Both streams are implemented as ConvNets.
Decoupling the spatial and temporal nets also allows us to
exploit the availability of large amounts of annotated image data by
pre-training the spatial net on the ImageNet challenge
dataset~\cite{Berg10a}.  Our proposed architecture is related
to the two-streams
hypothesis~\cite{Goodale92}, according to which the human visual
cortex contains two pathways: the ventral stream (which performs
object recognition) and the dorsal stream (which recognises
motion); though we do not investigate this connection any further here.

The rest of the paper is organised as
follows. In~\sref{sec:related_work} we review the related work on
action recognition using both shallow and deep architectures. In~\sref{sec:arch}
we introduce the two-stream architecture and specify the Spatial ConvNet. 
\sref{sec:temp_net} introduces the Temporal ConvNet and in particular how
it generalizes the previous architectures reviewed in~\sref{sec:related_work}.
A mult-task learning framework is developed in~\sref{sec:multi} in
order to allow effortless combination of training data over multiple datasets.
Implementation details are given in~\sref{sec:impl}, and the performance is evaluated
in~\sref{sec:eval} and compared to the state of the art.
Our experiments on two challenging datasets
(UCF-101~\cite{Soomro12} and HMDB-51~\cite{Kuehne11}) show that the
two recognition streams are complementary, and our deep architecture
significantly outperforms that of~\cite{Karpathy14} and is competitive
with the state of the art shallow
representations~\cite{Wang13b,Peng14,Peng14a} in spite of being trained on
relatively small datasets.

\subsection{Related work}
\label{sec:related_work}
Video recognition research has been largely driven by the advances in image recognition methods, which were often adapted and extended to deal with video data.
A large family of video action recognition methods is based on shallow high-dimensional encodings of local spatio-temporal features. 
For instance, the algorithm of~\cite{Laptev08} consists in detecting sparse spatio-temporal interest points,
which are then described using local spatio-temporal features: Histogram of Oriented Gradients (HOG)~\cite{Dalal05} and Histogram of Optical Flow (HOF). The features
are then encoded into the Bag Of Features (BoF) representation, which is pooled over several spatio-temporal grids (similarly to spatial pyramid pooling)
and combined with an SVM classifier. In a later work~\cite{Wang09}, it was shown that dense sampling of local features outperforms sparse interest points.

Instead of computing local video features over spatio-temporal cuboids, state-of-the-art shallow video representations~\cite{Wang13b,Peng14,Peng14a} make use of dense point trajectories.
The approach, first introduced in~\cite{Wang11b}, consists in adjusting local descriptor support regions, so that they follow dense trajectories, computed using optical flow.
The best performance in the trajectory-based pipeline was achieved by the Motion Boundary Histogram (MBH)~\cite{Dalal06}, which is a
gradient-based feature, separately computed on the horizontal and vertical components of optical flow. A combination of several features was shown to further boost the accuracy.
Recent improvements of trajectory-based hand-crafted representations include compensation of global (camera) motion~\cite{Jain13,Kuehne11,Wang13b}, and the use of the Fisher vector 
encoding~\cite{Perronnin10a} (in~\cite{Wang13b}) or its deeper variant~\cite{Simonyan13b} (in~\cite{Peng14a}).

There has also been a number of attempts to develop a deep architecture for video recognition. In the majority of these works, the input to the network is a stack of consecutive video
frames, so the model is expected to implicitly learn spatio-temporal motion-dependent features in the first layers, which can be a difficult task.
In~\cite{Jhuang07}, an HMAX architecture for video recognition was proposed with pre-defined spatio-temporal filters in the first layer.
Later, it was combined~\cite{Kuehne11} with a spatial HMAX model, thus forming spatial (ventral-like) and temporal (dorsal-like) recognition streams. 
Unlike our work, however, the streams were implemented as hand-crafted and rather shallow (3-layer) HMAX models.
In~\cite{Taylor10,Chen10,Le11}, a convolutional RBM and ISA were used for unsupervised learning of spatio-temporal features, which were then plugged into a discriminative model for action classification.
Discriminative end-to-end learning of video ConvNets has been addressed in~\cite{Ji13} and, more recently, in~\cite{Karpathy14}, who compared several ConvNet architectures for action recognition.
Training was carried out on a very large Sports-1M dataset, comprising 1.1M YouTube videos of sports activities.
Interestingly,~\cite{Karpathy14} found that a network, operating on individual video frames, performs similarly to the networks, whose input is a stack of frames.
This might indicate that the learnt spatio-temporal features do not capture the motion well.
The learnt representation, fine-tuned on the UCF-101 dataset, turned out to be $20\%$ less accurate than hand-crafted state-of-the-art trajectory-based representation~\cite{Wang13c,Peng14}. 

Our temporal stream ConvNet operates on multiple-frame dense optical flow, which
is typically computed in an energy minimisation framework by solving for a displacement field (typically at multiple image scales). 
We used a popular method of~\cite{Brox04}, which formulates the energy based on constancy assumptions for intensity and its gradient, as well as smoothness of the
displacement field. Recently,~\cite{Weinzaepfel13} proposed an image patch matching scheme, which is reminiscent of deep ConvNets, but does not incorporate learning.

\section{Two-stream architecture for video recognition}
\label{sec:arch}
\begin{figure}[ht]
\centering
\includegraphics[width=.9\textwidth]{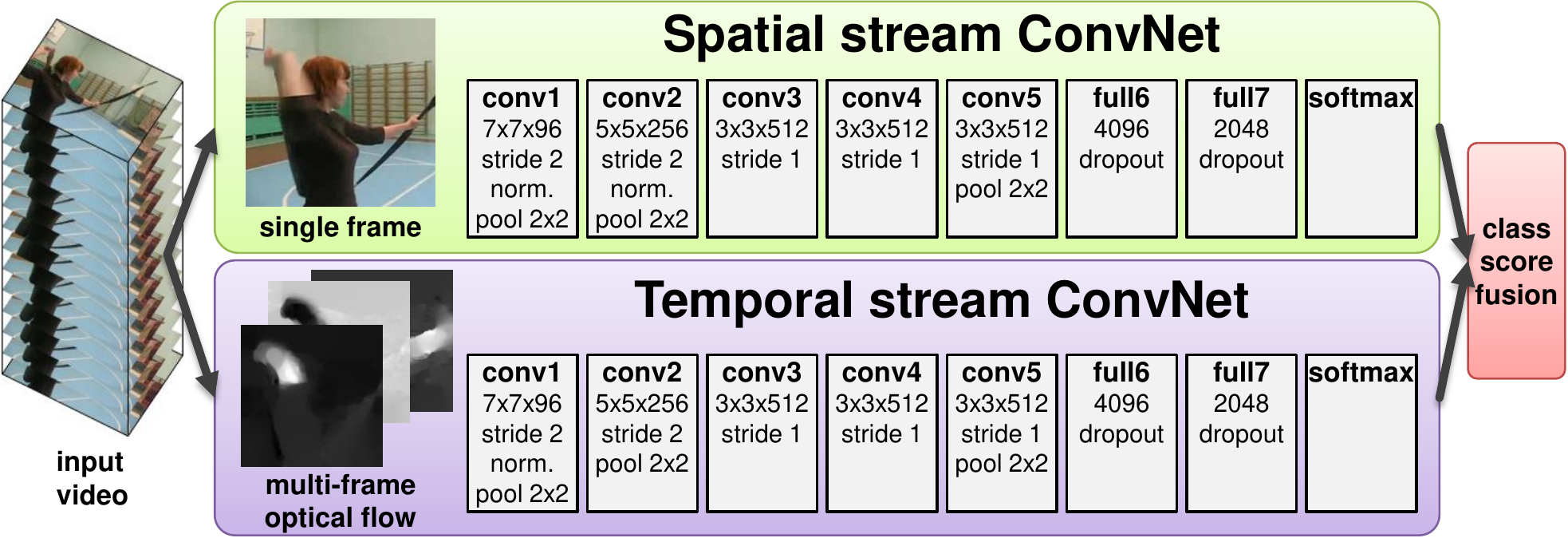}
\caption{\textbf{Two-stream architecture for video classification.}}
\label{fig:arch}
\end{figure}

Video can naturally be decomposed into spatial and temporal components. The spatial part, in the form of individual frame appearance, carries information about scenes and 
objects depicted in the video. The temporal part, in the form of motion across the frames, conveys the movement of the observer (the camera) and the objects. 
We devise our video recognition architecture accordingly, dividing it into two streams, as shown in~\figref{fig:arch}. Each stream is implemented using a deep ConvNet, 
softmax scores of which are combined by late fusion. We consider two fusion methods: averaging and training a multi-class linear SVM~\cite{Crammer01} on stacked $L_2$-normalised softmax scores
as features.

\noindent\textbf{Spatial stream ConvNet} operates on individual video frames, effectively performing action recognition from still images.
The static appearance by itself is a useful clue, since some actions are strongly associated with particular objects. 
In fact, as will be shown in~\sref{sec:eval}, action classification from still frames (the spatial recognition stream) is fairly competitive on its own.
Since a spatial ConvNet is essentially an image classification architecture, we can build upon the recent advances in large-scale image recognition methods~\cite{Krizhevsky12},
and pre-train the network on a large image classification dataset, such as the ImageNet challenge dataset. The details are presented in~\sref{sec:impl}.
Next, we describe the temporal stream ConvNet, which exploits motion and significantly improves accuracy.

\section{Optical flow ConvNets}
\label{sec:temp_net}
In this section, we describe a ConvNet model, which forms the temporal recognition stream of our architecture (\sref{sec:arch}).
Unlike the ConvNet models, reviewed in~\sref{sec:related_work}, the input to our model is formed by stacking optical flow displacement fields
between several consecutive frames. Such input explicitly describes the motion between video frames, which makes the recognition easier, as the network does 
not need to estimate motion implicitly. We consider several variations of the optical flow-based input, which we describe below. 
\begin{figure}[ht]
\centering
\begin{subfigure}[b]{0.19\textwidth}
\includegraphics[width=\textwidth]{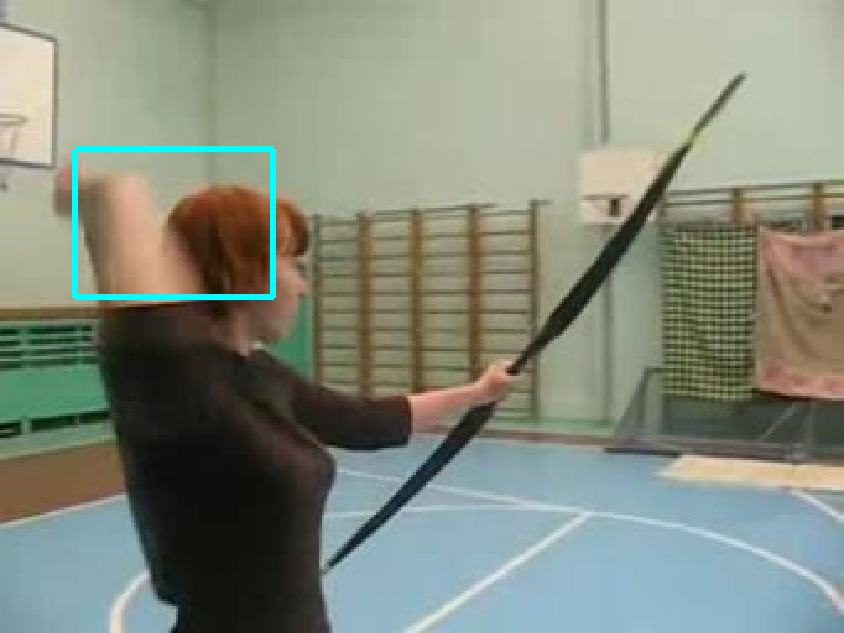}
\caption{}
\end{subfigure}
\begin{subfigure}[b]{0.19\textwidth}
\includegraphics[width=\textwidth]{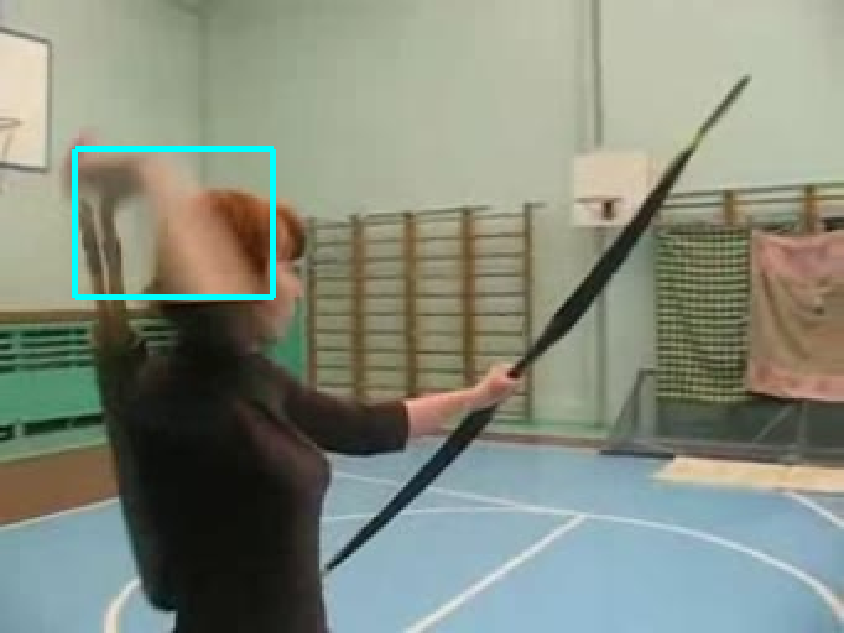}
\caption{}
\end{subfigure}
\begin{subfigure}[b]{0.19\textwidth}
\includegraphics[width=\textwidth]{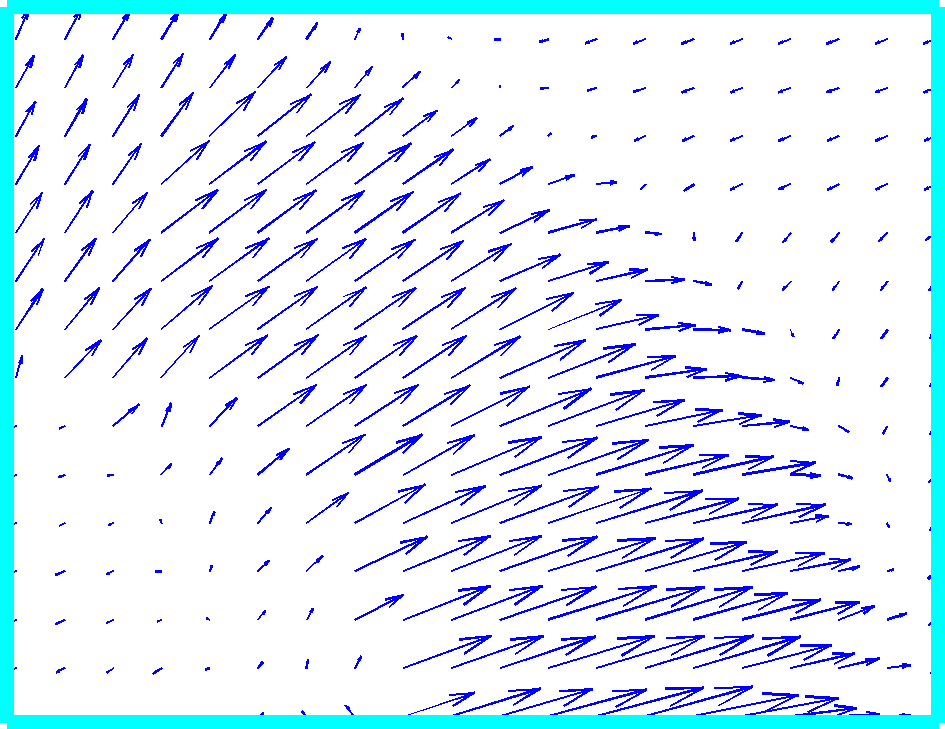}
\caption{}
\end{subfigure}
\begin{subfigure}[b]{0.19\textwidth}
\includegraphics[width=\textwidth]{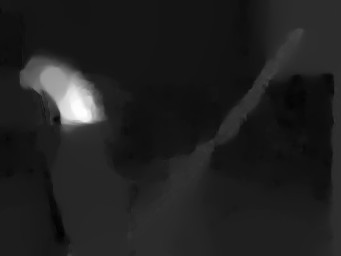}
\caption{}
\end{subfigure}
\begin{subfigure}[b]{0.19\textwidth}
\includegraphics[width=\textwidth]{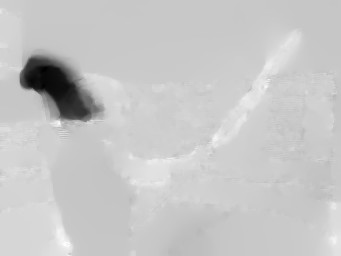}
\caption{}
\end{subfigure}
\caption{\textbf{Optical flow.} 
(a),(b): a pair of consecutive video frames with the area around a moving hand outlined with a cyan rectangle.
(c): a close-up of dense optical flow in the outlined area; 
(d): horizontal component $d^x$ of the displacement vector field (higher intensity corresponds to positive values, lower intensity to negative values).
(e): vertical component $d^y$.
Note how (d) and (e) highlight the moving hand and bow. The input to a ConvNet contains multiple flows (\sref{sec:temp_net_input}).
}
\label{fig:flow_vis}
\end{figure}

\subsection{ConvNet input configurations}
\label{sec:temp_net_input}

\begin{figure}[ht]
\centering
\includegraphics[width=\textwidth]{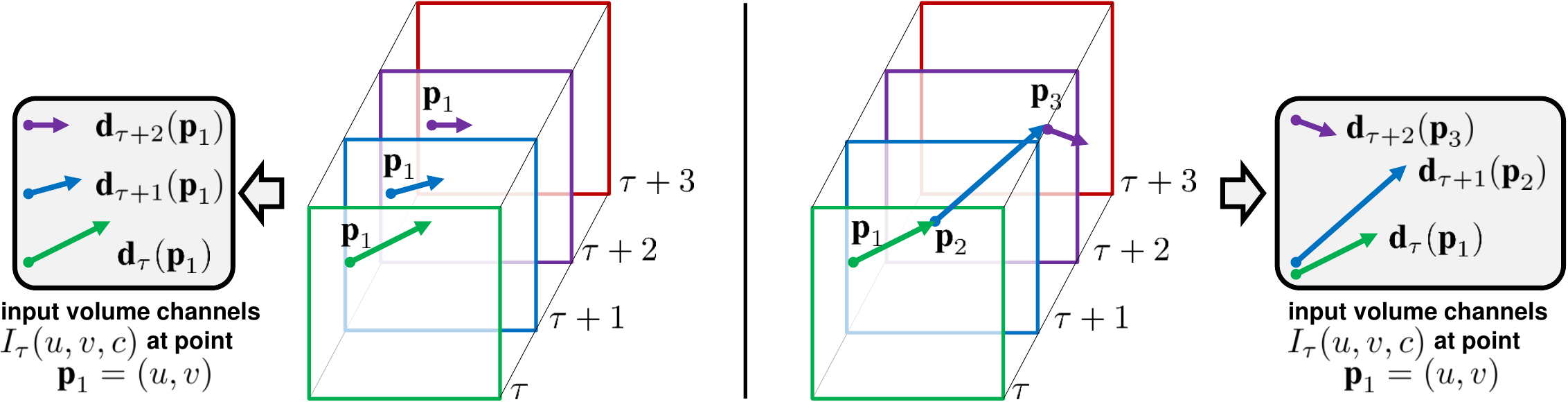}
\caption{\textbf{ConvNet input derivation from the multi-frame optical flow.}
\emph{Left:} optical flow stacking~\eqref{eq:input_flow} samples the displacement vectors $\dvec$ at the same location in multiple frames.
\emph{Right:} trajectory stacking~\eqref{eq:input_traj} samples the vectors along the trajectory. The frames and the
corresponding displacement vectors are shown with the same colour.
}
\label{fig:traj}
\end{figure}

\noindent\textbf{Optical flow stacking.}
A dense optical flow can be seen as a set of displacement vector fields $\dvec_t$ between the pairs of consecutive frames $t$ and $t+1$. 
By $\dvec_t(u,v)$ we denote the displacement vector at the point $(u,v)$ in frame $t$, which moves the point to the corresponding point in the following frame $t+1$. 
The horizontal and vertical components of the vector field, $d^x_t$ and $d^y_t$,  can be seen as image channels (shown in~\figref{fig:flow_vis}), well suited to
recognition using a convolutional network. 
To represent the motion across a sequence of frames, we stack the flow channels $d^{x,y}_t$ of $L$ consecutive frames to form a total of $2L$ input channels. 
More formally, let $w$ and $h$ be the width and height of a video;
a ConvNet input volume $I_{\tau} \in \Rbb^{w \times h \times 2L}$ for an arbitrary frame $\tau$ is then constructed as follows:
\begin{align}
\label{eq:input_flow}
&I_{\tau}(u,v,2k-1) = d^x_{\tau+k-1}(u,v), \\
\nonumber&I_{\tau}(u,v,2k) = d^y_{\tau+k-1}(u,v), \quad u=[1;w], v=[1;h], k=[1;L]. 
\end{align}
For an arbitrary point $(u,v)$, the channels $I_{\tau}(u,v,c), c=[1;2L]$ encode the motion at that point over a sequence of $L$ frames (as illustrated in \figref{fig:traj}-left).

\noindent\textbf{Trajectory stacking.}
An alternative motion representation, inspired by the trajectory-based descriptors~\cite{Wang11b}, replaces the optical flow, sampled at the same locations across several frames,
with the flow, sampled along the motion trajectories. In this case, the input volume $I_{\tau}$, corresponding to a frame $\tau$, takes the following form:
\begin{align}
\label{eq:input_traj}
&I_{\tau}(u,v,2k-1) = d^x_{\tau+k-1}(\pvec_k), \\
\nonumber&I_{\tau}(u,v,2k) = d^y_{\tau+k-1}(\pvec_k), \quad u=[1;w], v=[1;h], k=[1;L].
\end{align}
where $\pvec_k$ is the $k$-th point along the trajectory, which starts at the location $(u,v)$ in the frame $\tau$ and is defined by the following recurrence relation:
\[
\pvec_1 = (u,v); ~~~~~
\pvec_k = \pvec_{k-1} + \dvec_{\tau+k-2}(\pvec_{k-1}),\; k > 1.
\]
Compared to the input volume representation~\eqref{eq:input_flow},
where the channels $I_{\tau}(u,v,c)$ store the displacement vectors at
the locations $(u,v)$, the input volume~\eqref{eq:input_traj} stores
the vectors sampled at the locations $\pvec_k$ along the trajectory
(as illustrated in~\figref{fig:traj}-right).

\noindent\textbf{Bi-directional optical flow.}
Optical flow representations~\eqref{eq:input_flow} and~\eqref{eq:input_traj} deal with the forward optical flow, \ie the displacement field $\dvec_t$ of the frame $t$ specifies the location of its pixels
in the following frame $t+1$. It is natural to consider an extension to a bi-directional optical flow, which can be obtained by computing an additional set of displacement fields in the opposite direction.
We then construct an input volume $I_{\tau}$ by stacking $L/2$ forward flows between frames $\tau$ and $\tau+L/2$ and $L/2$ backward flows between frames $\tau-L/2$ and $\tau$. The input $I_{\tau}$ 
thus has the same number of channels ($2L$) as before. The flows can be represented using either of the two methods~\eqref{eq:input_flow} and~\eqref{eq:input_traj}.

\noindent\textbf{Mean flow subtraction.}
It is generally beneficial to perform zero-centering of the network input, as it allows the model to better exploit the rectification non-linearities.
In our case, the displacement vector field components can take on both positive and negative values, and are naturally centered in the sense that across a large variety of motions,
the movement in one direction is as probable as the movement in the opposite one. However, given a pair of frames, the optical flow between them can be dominated by a particular
displacement, \eg caused by the camera movement. 
The importance of camera motion compensation has been previously highlighted in~\cite{Jain13,Wang13b}, where a global motion component was estimated and subtracted from the dense flow.
In our case, we consider a simpler approach: from each displacement field $\dvec$ we subtract its mean vector. 

\noindent\textbf{Architecture.} Above we have described different ways of combining multiple optical flow displacement fields into a single volume $I_{\tau} \in \Rbb^{w \times h \times 2L}$.
Considering that a ConvNet requires a fixed-size input, we sample a $224\times224\times2L$ sub-volume from $I_{\tau}$ and pass it to the net as input.
The hidden layers configuration remains largely the same as 
that used in the spatial net, and is illustrated in~\figref{fig:arch}.
Testing is similar to the spatial ConvNet, and is described in detail in~\sref{sec:impl:temporal}.

\subsection{Relation of the temporal ConvNet architecture to previous representations}
\label{sec:rel}

In this section, we put our temporal ConvNet architecture in the
context of prior art, drawing connections to the video
representations, reviewed in~\sref{sec:related_work}.  Methods based
on feature encodings~\cite{Laptev08,Wang11b} typically combine several
spatio-temporal local features. Such features are computed from the
optical flow and are thus generalised by our temporal ConvNet.
Indeed, the HOF and MBH local descriptors are based on the
histograms of orientations of optical flow or its gradient, which can be obtained from
the displacement field input~\eqref{eq:input_flow} using a single
convolutional layer (containing orientation-sensitive filters),
followed by the rectification and pooling layers. 
The kinematic features of~\cite{Jain13} (divergence, curl and shear)
are also computed from the optical flow gradient, and, again, can be captured
by our convolutional model.  Finally, the trajectory
feature~\cite{Wang11b} is computed by stacking the displacement
vectors along the trajectory, which corresponds to the
trajectory stacking~\eqref{eq:input_traj}.  
In~\sref{sec:vis_conv} we visualise the convolutional filters, learnt in the first layer of the temporal network. 
This provides further evidence that our representation generalises hand-crafted features.

As far as the deep networks are concerned, a two-stream video classification architecture of~\cite{Kuehne11} contains two HMAX models which are hand-crafted and less deep than our
discriminatively trained ConvNets, which can be seen as a learnable generalisation of HMAX. The convolutional models of~\cite{Ji13,Karpathy14} do not decouple spatial and temporal
recognition streams, and rely on the motion-sensitive convolutional filters, learnt from the data. In our case, motion is explicitly represented using the optical flow displacement
field, computed based on the assumptions of constancy of the intensity and smoothness of the flow. Incorporating such assumptions into a ConvNet framework might be able to boost the 
performance of end-to-end ConvNet-based methods, and is an interesting direction for future research.

\subsection{Visualisation of learnt convolutional filters}
\label{sec:vis_conv}

\begin{figure}[ht]
\centering
\includegraphics[width=\textwidth]{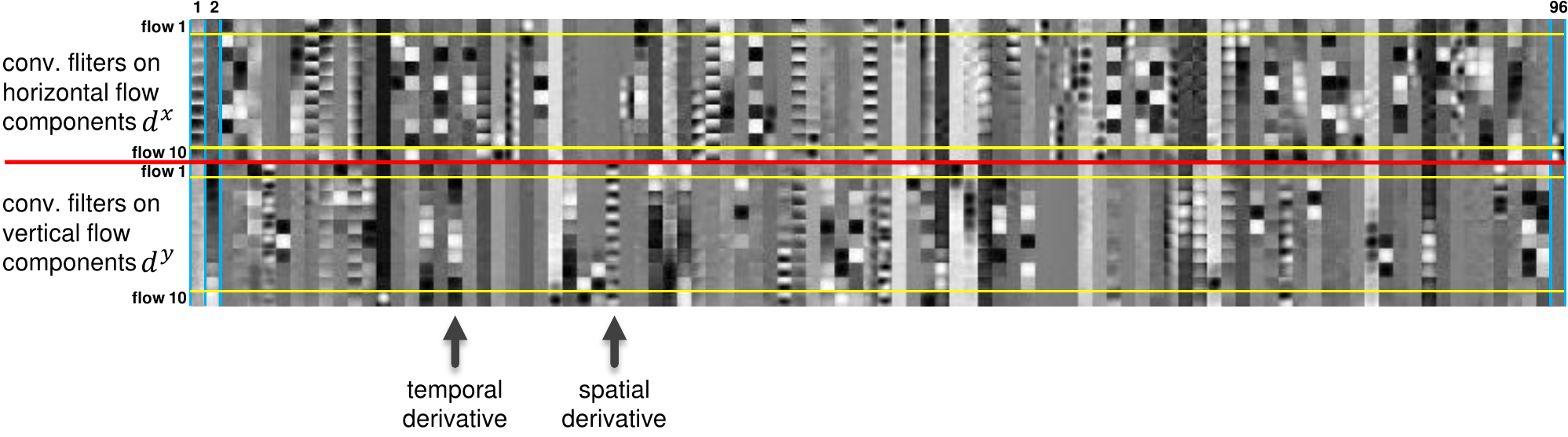}
\caption{\textbf{First-layer convolutional filters learnt on 10 stacked optical flows.}
The visualisation is split into 96 columns and 20 rows: each column corresponds to a filter, each row -- to an input channel.}
\label{fig:conv1}
\end{figure}
In~\figref{fig:conv1} we visualise the convolutional filters from the first layer of the temporal ConvNet, trained on the UCF-101 dataset.
Each of the $96$ filters has a spatial receptive field of $7 \times 7$ pixels, and spans 20 input channels, corresponding to the horizontal ($d^x$)
and vertical ($d^y$) components of $10$ stacked optical flow displacement fields $\dvec$.

As can be seen, some filters compute spatial derivatives of the optical flow, capturing how motion changes with image location, which generalises
derivative-based hand-crafted descriptors (\eg MBH).
Other filters compute temporal derivatives, capturing changes in motion over time.

\section{Multi-task learning}
\label{sec:multi}

Unlike the spatial stream ConvNet, which can be pre-trained on a large
still image classification dataset (such as ImageNet), the temporal
ConvNet needs to be trained on video data -- and the available datasets
for video action classification are still rather small.  In our experiments
(\sref{sec:eval}), training is performed on the UCF-101 and HMDB-51
datasets, which have only: 9.5K and 3.7K videos
respectively.  To decrease over-fitting, one could consider combining
the two datasets into one; this, however, is not straightforward due
to the intersection between the sets of classes. 
One option (which we evaluate later) is to only add the images from the classes, which do not appear in the original dataset.
This, however, requires manual search for such classes and limits the amount of additional training data.

A more principled way of combining several datasets is based on multi-task learning~\cite{Collobert08}.
Its aim is to learn a (video) representation, which is applicable not only to the task in question (such as HMDB-51 classification), but also
to other tasks (\eg UCF-101 classification). Additional tasks act as a regulariser, and allow for the exploitation of additional training data.
In our case, a ConvNet architecture is modified so that it has \emph{two} softmax classification layers on
top of the last fully-connected layer: one softmax layer computes HMDB-51 classification scores, the other one -- the UCF-101 scores. Each of the layers is equipped with
its own loss function, which operates only on the videos, coming from the respective dataset. The overall training loss is computed as the sum of the individual
tasks' losses, and the network weight derivatives can be found by back-propagation.



\section{Implementation details}
\label{sec:impl}
\label{sec:impl:spatial}
\label{sec:impl:temporal}

\noindent\textbf{ConvNets configuration.}
The layer configuration of our spatial and temporal ConvNets is schematically shown in~\figref{fig:arch}.
It corresponds to CNN-M-2048 architecture of~\cite{Chatfield14} and is similar to the network of~\cite{Zeiler13}.
All hidden weight layers use the rectification (ReLU) activation function; max-pooling is performed over $3\times 3$ spatial windows with stride 2; local response
normalisation uses the same settings as~\cite{Krizhevsky12}.
The only difference between spatial and temporal ConvNet configurations is that we removed the second normalisation layer from the latter to reduce memory consumption. 

\noindent\textbf{Training.} The training procedure can be seen as an adaptation of that of~\cite{Krizhevsky12} to video frames, and is generally the same for both spatial and temporal nets.
The network weights are learnt using the mini-batch stochastic gradient descent with momentum (set to 0.9).
At each iteration, a mini-batch of 256 samples is constructed by sampling 256 training videos (uniformly across the classes), from each of which a single frame is randomly selected.
In spatial net training,  a $224 \times 224$ sub-image is randomly cropped from the selected frame; it then undergoes random horizontal flipping and RGB jittering.
The videos are rescaled beforehand, so that the smallest side of the frame equals 256. 
We note that unlike~\cite{Krizhevsky12}, the sub-image is sampled from the whole frame, not just its $256 \times 256$ center.
In the temporal net training, we compute an optical flow volume $I$ for the selected training frame as described in~\sref{sec:temp_net}.
From that volume, a fixed-size $224\times224\times2L$ input is randomly cropped and flipped.
The learning rate is initially set to $10^{-2}$, and then decreased according to a fixed schedule, which is kept the same for all training sets. 
Namely, when training a ConvNet from scratch, the rate is changed to $10^{-3}$ after 50K iterations, then to $10^{-4}$ after 70K iterations, and training is stopped after 80K iterations.
In the fine-tuning scenario, the rate is changed to $10^{-3}$ after 14K iterations, and training stopped after 20K iterations.

\noindent\textbf{Testing.} At test time, given a video, we sample a fixed number of frames ($25$ in our experiments) with equal temporal spacing between them. 
From each of the frames we then obtain 10 ConvNet inputs~\cite{Krizhevsky12} by cropping and flipping four corners and the center of the frame.
The class scores for the whole video are then obtained by averaging the scores across the sampled frames and crops therein.

\noindent\textbf{Pre-training on ImageNet ILSVRC-2012.}
When pre-training the spatial ConvNet, we use the same training and test data augmentation as described above (cropping, flipping, RGB jittering). 
This yields $13.5\%$ top-5 error on ILSVRC-2012 validation set, which compares favourably to $16.0\%$ reported in~\cite{Zeiler13} for a similar network. 
We believe that the main reason for the improvement is sampling of ConvNet inputs from the whole image, rather than just its center.

\noindent\textbf{Multi-GPU training.}
Our implementation is derived from the publicly available Caffe toolbox~\cite{Jia13}, but contains a number of significant modifications, including parallel 
training on multiple GPUs installed in a single system. We exploit the data parallelism, and split each SGD batch across several GPUs.
Training a single temporal ConvNet takes 1 day on a system with 4 NVIDIA Titan cards, which constitutes a $3.2$ times speed-up over single-GPU training.

\noindent\textbf{Optical flow} is computed using the off-the-shelf GPU implementation of~\cite{Brox04} from the OpenCV toolbox.
In spite of the fast computation time ($0.06$s for a pair of frames), it would still introduce a bottleneck if done on-the-fly, 
so we pre-computed the flow before training. To avoid storing the displacement fields as floats, the horizontal and vertical components of the flow were 
linearly rescaled to a $[0,255]$ range and compressed using JPEG (after decompression, the flow is rescaled back to its original range).
This reduced the flow size for the UCF-101 dataset from 1.5TB to 27GB.

\section{Evaluation}
\label{sec:eval}
\noindent\textbf{Datasets and evaluation protocol.}
The evaluation is performed on UCF-101~\cite{Soomro12} and \mbox{HMDB-51}~\cite{Kuehne11} action recognition benchmarks, which are among the largest available annotated video
datasets\footnote{Very recently,~\cite{Karpathy14} released the Sports-1M dataset of 1.1M automatically annotated YouTube sports videos. 
Processing the dataset of such scale is very challenging, and we plan to address it in future work.}. 
UCF-101 contains 13K videos (180 frames/video on average), annotated into 101 action classes; HMDB-51 includes 6.8K videos of 51 actions. 
The evaluation protocol is the same for both datasets: the organisers
provide three splits into training and test data, and the performance is measured by the mean classification accuracy across the splits. Each UCF-101 split contains 9.5K training
videos; an HMDB-51 split contains 3.7K training videos.
We begin by comparing different architectures on the first split of the UCF-101 dataset.
For comparison with the state of the art, we follow the standard evaluation protocol and report the average accuracy over three splits on both UCF-101 and HMDB-51. 

\noindent\textbf{Spatial ConvNets.}
First, we measure the performance of the spatial stream ConvNet.
Three scenarios are considered:
(i) training from scratch on UCF-101, 
(ii) pre-training on ILSVRC-2012 followed by fine-tuning on UCF-101, 
(iii) keeping the pre-trained network fixed and only training the last (classification) layer.  
For each of the settings, we experiment with setting the dropout
regularisation ratio to $0.5$ or to $0.9$.  
From the results,
presented in~\tblref{tab:spatial}, it is clear that training the
ConvNet solely on the UCF-101 dataset leads to over-fitting (even with
high dropout), and is inferior to pre-training on a large ILSVRC-2012
dataset. Interestingly, fine-tuning the whole network gives only
marginal improvement over training the last layer only.
In the latter setting, higher dropout over-regularises learning and leads to worse accuracy. 
In the following experiments we opted for training the last layer on top of a pre-trained ConvNet.
\begin{table}[ht]
\setlength{\tabcolsep}{4pt}
\small
\caption{\textbf{Individual ConvNets accuracy on UCF-101 (split 1).}}
\begin{subtable}{.4\textwidth}
\vspace{-2em}
\centering
\caption{\textbf{Spatial ConvNet.}}
\begin{tabular}{|l|c|c|} \hline
\multirow{2}{*}{Training setting} & \multicolumn{2}{c|}{Dropout ratio} \\ \cline{2-3}
 & $0.5$ & $0.9$  \\ \hline
From scratch & 42.5\%  & 52.3\% \\ \hline
Pre-trained + fine-tuning & 70.8\%  & \textbf{72.8\%}  \\ \hline
Pre-trained + last layer & \textbf{72.7\%} & 59.9\%  \\ \hline
\end{tabular}
\label{tab:spatial}
\end{subtable}
~~~
\begin{subtable}{.6\textwidth}
\centering
\caption{\textbf{Temporal ConvNet.}}
\begin{tabular}{|l|c|c|} \hline
\multirow{2}{*}{Input configuration} & \multicolumn{2}{c|}{Mean subtraction} \\ \cline{2-3}
 & off & on  \\ \hline
Single-frame optical flow ($L=1$) & - & 73.9\% \\ \hline
Optical flow stacking~\eqref{eq:input_flow} ($L=5$) & - & 80.4\% \\ \hline
Optical flow stacking~\eqref{eq:input_flow} ($L=10$) & 79.9\% & \textbf{81.0\%} \\ \hline
Trajectory stacking~\eqref{eq:input_traj}($L=10$) & 79.6\% & 80.2\% \\ \hline
Optical flow stacking~\eqref{eq:input_flow}($L=10$), bi-dir.  & - & \textbf{81.2\%} \\ \hline
\end{tabular}
\label{tab:temporal}
\end{subtable}
\end{table}

\noindent\textbf{Temporal ConvNets.}
Having evaluated spatial ConvNet variants, we now turn to the temporal ConvNet architectures, and assess the effect of the input configurations, described in~\sref{sec:temp_net_input}.
In particular, we measure the effect of: using multiple ($L=\{5,10\}$) stacked optical flows; trajectory stacking; mean displacement subtraction; using the bi-directional optical flow.
The architectures are trained on the UCF-101 dataset from scratch, so we used an aggressive dropout ratio of $0.9$ to help improve generalisation. 
The results are shown in~\tblref{tab:temporal}.
First, we can conclude that stacking multiple ($L>1$) displacement fields in the input is highly beneficial, as it provides the network with long-term motion information, which is more discriminative 
than the flow between a pair of frames ($L=1$ setting). Increasing the number of input flows from $5$ to $10$ leads to a smaller improvement, so we kept $L$ fixed to $10$ in the following experiments.
Second, we find that mean subtraction is helpful, as it reduces the effect of global motion between the frames. We use it in the following experiments as default.
The difference between different stacking techniques is marginal; it turns out that optical flow stacking performs better than trajectory stacking,
and using the bi-directional optical flow is only slightly better than a uni-directional forward flow.
Finally, we note that temporal ConvNets significantly outperform the spatial ConvNets (\tblref{tab:spatial}), which confirms the importance of motion information for action recognition.

We also implemented the ``slow fusion'' architecture of~\cite{Karpathy14}, which amounts to applying a \mbox{ConvNet} to a stack of RGB frames ($11$ frames in our case).
When trained from scratch on UCF-101, it achieved $56.4\%$ accuracy, which is better than a single-frame architecture trained from scratch ($52.3\%$), but is still far off the network 
trained from scratch on optical flow. This shows that while multi-frame information is important, it is also important to present it to a ConvNet in an appropriate manner.

\noindent\textbf{Multi-task learning of temporal ConvNets.}
Training temporal ConvNets on UCF-101 is challenging due to the small size of the training set. An even bigger challenge is to train the ConvNet on HMDB-51, where
each training split is $2.6$ times smaller than that of UCF-101. Here we evaluate different options for increasing the effective training set size of HMDB-51:
(i) fine-tuning a temporal network pre-trained on UCF-101; 
(ii) adding $78$ classes from UCF-101, which are manually selected so that there is no intersection between
these classes and the native HMDB-51 classes; 
(iii) using the multi-task formulation (\sref{sec:multi}) to learn a video representation, shared between the UCF-101 and HMDB-51 classification
tasks. The results are reported in~\tblref{tab:hmdb_extra_data}. 
\begin{table}[ht]
\small
\centering
\caption{\textbf{Temporal ConvNet accuracy on HMDB-51 (split 1 with additional training data).}
}
\begin{tabular}{|l|c|} \hline
\multicolumn{1}{|l|}{Training setting} & Accuracy \\ \hline
Training on HMDB-51 without additional data & 46.6\%  \\ \hline
Fine-tuning a ConvNet, pre-trained on UCF-101 & 49.0\%  \\ \hline
Training on HMDB-51 with classes added from UCF-101 & 52.8\% \\ \hline
Multi-task learning on HMDB-51 and UCF-101 & \textbf{55.4\%} \\ \hline
\end{tabular}
\label{tab:hmdb_extra_data}
\end{table}
As expected, it is beneficial to utilise full (all splits combined) UCF-101 data for training (either explicitly by borrowing images, or implicitly by pre-training). 
Multi-task learning performs the best, as it allows the training procedure to exploit all available training data.

We have also experimented with multi-task learning on the UCF-101 dataset, by training a network to classify both the full HMDB-51 data (all splits combined) and
the UCF-101 data (a single split). On the first split of UCF-101, the accuracy was measured to be 81.5\%,
which improves on $81.0\%$ achieved using the same settings, but without the additional HMDB classification task (\tblref{tab:temporal}).

\noindent\textbf{Two-stream ConvNets.}
Here we evaluate the complete two-stream model, which combines the two recognition streams. 
One way of combining the networks would be to train a joint stack of fully-connected layers on top of full6 or full7 layers of the two nets.
This, however, was not feasible in our case due to over-fitting. We therefore fused the softmax scores using either averaging or a linear SVM.
From~\tblref{tab:fusion} we conclude that: 
(i)~\emph{temporal and spatial recognition streams are complementary, as their fusion significantly improves on both} ($6\%$ over temporal and $14\%$ over spatial nets);
(ii)~SVM-based fusion of softmax scores outperforms fusion by averaging;
(iii)~using bi-directional flow is not beneficial in the case of ConvNet fusion; 
(iv)~temporal ConvNet, trained using multi-task learning, performs the best both alone and when fused with a spatial net.
\begin{table}[ht]
\small
\centering
\caption{\textbf{Two-stream ConvNet accuracy on UCF-101 (split 1).}
}
\begin{tabular}{|l|l|l|c|} \hline
Spatial ConvNet & Temporal ConvNet & Fusion Method & Accuracy \\ \hline
Pre-trained + last layer & bi-directional & averaging & 85.6\% \\ \hline
Pre-trained + last layer & uni-directional & averaging & 85.9\%  \\ \hline
Pre-trained + last layer & uni-directional, multi-task & averaging  & 86.2\% \\ \hline
Pre-trained + last layer & uni-directional, multi-task & SVM & \textbf{87.0\%} \\ \hline
\end{tabular}
\label{tab:fusion}
\end{table}

\noindent\textbf{Comparison with the state of the art.}
We conclude the experimental evaluation with the comparison against the state of the art on three splits of UCF-101 and HMDB-51.
For that we used a spatial net, pre-trained on ILSVRC, with the last layer trained on UCF or HMDB. 
The temporal net was trained on UCF and HMDB using multi-task learning, and the input was computed using uni-directional optical flow stacking with 
mean subtraction. The softmax scores of the two nets were combined using averaging or SVM.
As can be seen from~\tblref{tab:comp_SOA}, both our spatial and temporal nets alone outperform the deep architectures of~\cite{Karpathy14,Kuehne11} by a large margin.
The combination of the two nets further improves the results (in line with the single-split experiments above), and is comparable to the very recent 
state-of-the-art hand-crafted models~\cite{Wang13b,Peng14,Peng14a}.

\begin{table}[ht]
\small
\centering
\caption{\textbf{Mean accuracy (over three splits) on UCF-101 and HMDB-51.}
}
\begin{tabular}{|l|c|c|} \hline
\multicolumn{1}{|c|}{Method} & UCF-101 & HMDB-51 \\ \hline
Improved dense trajectories (IDT)~\cite{Wang13b,Wang13c} & 85.9\% & 57.2\% \\ \hline
IDT with higher-dimensional encodings~\cite{Peng14} & \textbf{87.9\%} & 61.1\% \\ \hline
IDT with stacked Fisher encoding~\cite{Peng14a} (based on Deep Fisher Net~\cite{Simonyan13b}) & - & \textbf{66.8\%} \\ \hline
Spatio-temporal HMAX network~\cite{Kuehne11,Jhuang07} & - & 22.8\% \\ \hline
``Slow fusion'' spatio-temporal ConvNet~\cite{Karpathy14} & 65.4\% & - \\ \hline\hline
Spatial stream ConvNet & 73.0\% & 40.5\% \\ \hline
Temporal stream ConvNet & 83.7\% & 54.6\% \\ \hline
Two-stream model (fusion by averaging) & 86.9\% & 58.0\% \\ \hline
Two-stream model (fusion by SVM) & \textbf{88.0\%} & \textbf{59.4\%} \\ \hline
\end{tabular}
\label{tab:comp_SOA}
\end{table}

\begin{figure}[htb]
\centering
\includegraphics[width=.8\textwidth]{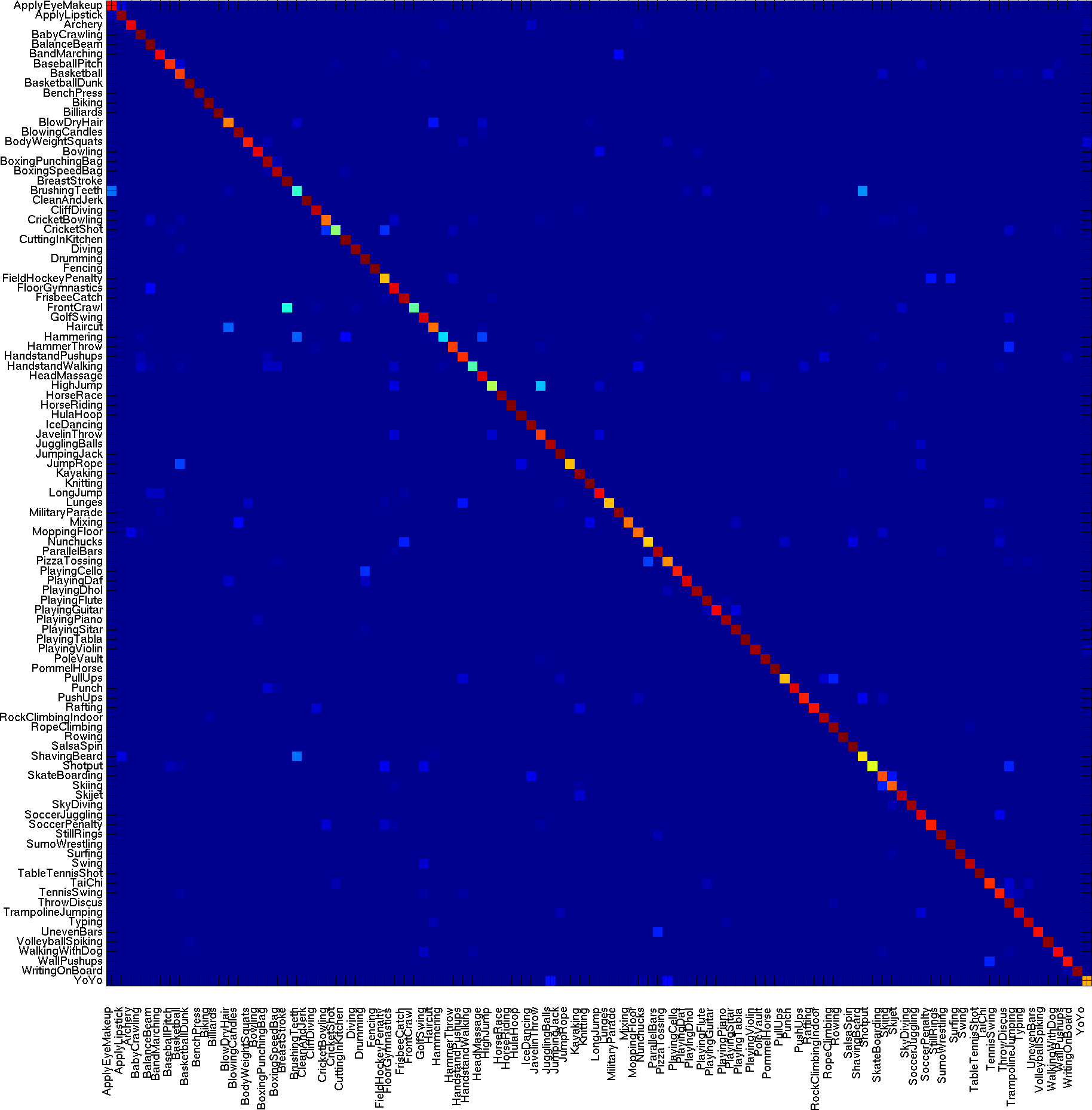}
\caption{\textbf{Confusion matrix of a two-stream model on the first split of UCF-101.} 
}
\label{fig:conf}
\end{figure}

\begin{figure}[htb]
\centering
\includegraphics[width=.8\textwidth]{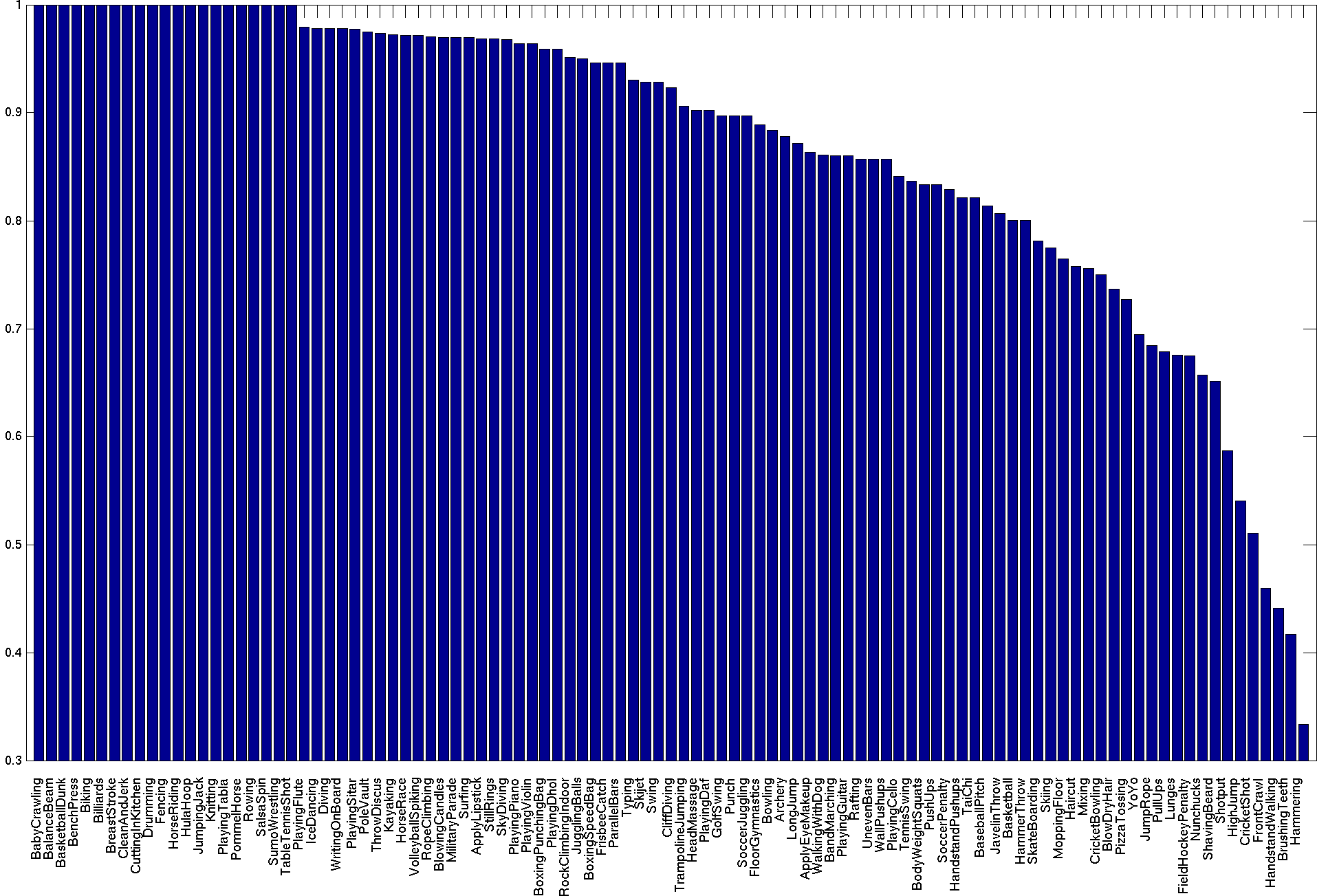}
\caption{\textbf{Per-class recall of a two-stream model on the first split of UCF-101.} 
}
\label{fig:recall}
\end{figure}

\noindent\textbf{Confusion matrix and per-class recall for UCF-101 classification.}
In~\figref{fig:conf} we show the confusion matrix for UCF-101 classification using our two-stream model, which achieves $87.0\%$ accuracy on the first dataset split (the last row of~\tblref{tab:fusion}).
We also visualise the corresponding per-class recall in~\figref{fig:recall}.

The worst class recall corresponds to \emph{Hammering} class, which is confused with \emph{HeadMassage} and \emph{BrushingTeeth} classes.
We found that this is due to two reasons. First, the spatial ConvNet confuses \emph{Hammering} with \emph{HeadMassage}, which can be caused by
the significant presence of human faces in both classes. Second, the temporal ConvNet confuses \emph{Hammering} with \emph{BrushingTeeth}, as
both actions contain recurring motion patterns (hand moving up and down).

\section{Conclusions and directions for improvement}
We proposed a deep video classification model with competitive performance,
which incorporates separate spatial and temporal recognition streams based on ConvNets.
Currently it appears that training a temporal ConvNet on optical flow
(as here) is significantly better than training on raw stacked
frames~\cite{Karpathy14}.  The latter is probably too challenging, and
might require architectural changes (for example, a combination with
the deep matching approach of~\cite{Weinzaepfel13}). Despite using
optical flow as input, our temporal model does not require significant
hand-crafting, since the flow is computed using a method based on the
generic assumptions of constancy and smoothness.

As we have shown, extra training data is beneficial for our temporal
ConvNet, so we are planning to train it on large video datasets,
such as the recently released collection of~\cite{Karpathy14}. This,
however, poses a significant challenge on its own due to the gigantic
amount of training data (multiple TBs).  

There still remain some essential ingredients of the state-of-the-art
shallow representation~\cite{Wang13b}, which are missed in our current
architecture.  The most prominent one is local feature pooling over
spatio-temporal tubes, centered at the trajectories. Even though the
input~\eqref{eq:input_traj} captures the optical flow along the
trajectories, the spatial pooling in our network does not take the
trajectories into account. Another potential area of improvement is
explicit handling of camera motion, which in our case is compensated
by mean displacement subtraction.

\section*{Acknowledgements}
This work was supported by ERC grant VisRec no. 228180. 
We gratefully acknowledge the support of NVIDIA Corporation with the donation of the GPUs used for this research.

\bibliographystyle{plainnat}
{
    \small
    \bibliography{bib/shortstrings,bib/vgg_local,bib/vgg_other,bib/current}
}

\end{document}